\documentclass{article}

\usepackage{PRIMEarxiv}

\usepackage[utf8]{inputenc} 
\usepackage[T1]{fontenc}    
\usepackage{hyperref}       
\usepackage{url}            
\usepackage{booktabs}       
\usepackage{amsfonts}       
\usepackage{nicefrac}       
\usepackage{microtype}      
\usepackage{lipsum}
\usepackage{fancyhdr}       
\usepackage{graphicx}       
\graphicspath{{media/}}     

\usepackage{microtype}
\usepackage{graphicx}
\usepackage{subcaption}
\usepackage{booktabs} 
\usepackage{tikz}\usetikzlibrary{decorations.pathreplacing,calligraphy}
\usepackage{dsfont}
\usepackage{multirow, makecell}
\usepackage{rotating}
\usepackage{booktabs}
\usepackage{hyperref}
\usepackage{threeparttable}
\usepackage{siunitx}
\usepackage[numbers]{natbib}

\usepackage{amsmath}
\usepackage{amssymb}
\usepackage{mathtools}
\usepackage{amsthm}
\usepackage{wrapfig}

\usepackage[capitalize,noabbrev]{cleveref}

\newcommand\BlueText{\color{blue}} 
\newcommand\RedText{\color{red}}
\title{Zero-Knowledge Zero-Shot Learning for Novel Visual Category Discovery
}

\author{
  Zhaonan Li \\
  Columbia University \\
  \texttt{zl3086@columbia.edu} \\
   \And
  Hongfu Liu \\
  Brandeis University \\
  \texttt{hongfuliu@brandeis.edu} \\
}

\begin{document}
\maketitle

\begin{abstract}
Generalized Zero-Shot Learning (GZSL) and Open-Set Recognition (OSR) are two mainstream settings that greatly extend conventional visual object recognition. However, the limitations of their problem settings are not negligible. The novel categories in GZSL require pre-defined semantic labels, making the problem setting less realistic; the oversimplified unknown class in OSR fails to explore the innate fine-grained and mixed structures of novel categories. In light of this, we are motivated to consider a new problem setting named {Zero-Knowledge Zero-Shot Learning} (ZK-ZSL) that assumes no prior knowledge of novel classes and aims to classify seen and unseen samples and recover semantic attributes of the fine-grained novel categories for further interpretation. To achieve this, we propose a novel framework that recovers the clustering structures of both seen and unseen categories where the seen class structures are guided by source labels. In addition, a structural alignment loss is designed to aid the semantic learning of unseen categories with their recovered structures. Experimental results demonstrate our method's superior performance in classification and semantic recovery on four benchmark datasets. 
\end{abstract}


\section{Introduction}
In recent years, visual recognition with deep learning has made tremendous progress. Closed-set is a common assumption in deep visual recognition where the training and testing datasets have a shared label space. However, data insufficiency and unavailability in real-world applications make it difficult to satisfy and pose significant challenges to visual classifiers. To this end, Generalized Zero-Shot Learning (GZSL) and Open-Set Recognition (OSR) are proposed to extend traditional visual recognition by acknowledging the existence of unknown categories in the target dataset. Specifically, GZSL aims to recognize both seen and unseen classes given labeled semantic attributes. The goal of OSR is to classify seen samples and detect unseen ones without prior information. Unfortunately, GZSL and OSR still have limitations in practical use. For example, the notion of novel\footnote{In this paper, we use unseen and novel interchangeably. } categories in GZSL is pre-determined by the provided semantic labels. However, knowing the novel semantic attributes implies the existence of images from unseen categories. Then the problem can be simply reduced to supervised learning by collecting training samples of novel categories. In this sense, the GZSL problem is less practical as the model is expected to automatically discover novel categories on its own. 
In addition, OSR recognizes unseen categories as a whole without further analysis of their intrinsic fine-grained structures. Furthermore, OSR does not consider the semantics of visual categories, failing to provide richer information for in-depth analyses and interpretation of novel categories.

\begin{wrapfigure}[15]{R}{0.5\textwidth}
    \vspace{-2mm}
    \centering
    \includegraphics[width=0.9 \linewidth]{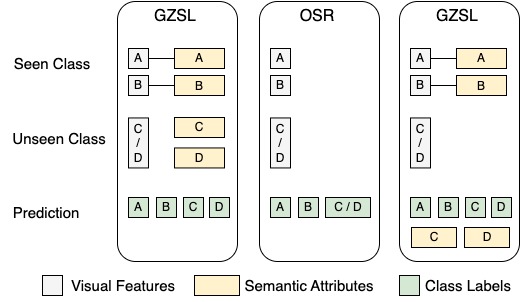}\vspace{-2mm}
    \caption{Schematic illustrations of GZSL, OSR, and our explored Zero-Knowledge Zero-Shot Learning (ZK-ZSL).}
    \label{fig:schemaicviews}
    \vspace{-5mm}
\label{sec:intro}
\end{wrapfigure}

In this paper, we explore a novel setting named Zero-Knowledge Zero-Shot Learning (ZK-ZSL) that addresses the aforementioned limitations of GZSL and OSR. Figure~\ref{fig:schemaicviews} illustrates the difference between GZSL, OSR, and our setting. ZK-ZSL aims to recognize both seen and unseen classes without prior knowledge of novel categories and recover their semantic attributes of the fine-grained novel categories. In this setting, only semantic labels of the seen categories are available. Compared with GZSL and OSR, the challenges of ZK-ZSL lie in the following aspects. (1) A classifier trained on the seen label space might separate seen samples from unseen samples, but cannot further separate samples in fine-grained unseen categories, and the biased embedding learned from the seen categories could potentially destroy underlying structures of unseen categories. (2) Since the semantic labels of unseen categories are unavailable, training only with seen semantics is susceptible to overfitting and leads to biased semantic predictions.

We propose a novel framework to tackle the ZK-ZSL problem and address the above challenges. In what follows, we first discuss how our proposed method takes into account the aforementioned challenges, and then we summarize our main contributions. To recognize seen classes and discover the structures of novel categories, our method learns to cluster on the target data consisting of both seen and unseen categories. Meanwhile, the class structures of seen categories are guided by the source labels. In addition, to mitigate the unavailability problem of unseen semantic labels, we enforce a structural alignment between recovered hidden embedding and predicted semantic space to aid the semantic learning of unseen categories with their discovered clustering structures. Our main contributions can be summarized as follows:
\begin{itemize}
    \vspace{-1mm}
    \item \parbox[t]{\dimexpr\textwidth-\leftmargin}{We formulate and explore a new visual recognition setting named Zero-Knowledge Zero-Shot Learning (ZK-ZSL) that aims to recognize seen and unseen categories without prior knowledge and recovers unseen semantic attributes, addressing the limitations of GZSL and OSR in practical use.}
    \vspace{-1mm}
    \item \parbox[t]{\dimexpr\textwidth-\leftmargin}{We propose a novel method that learns to uncover clustering structures in the target dataset guided by labeled seen class structures. A novel structural alignment loss is designed to aid the learning of novel semantic attributes with recovered clustering structures.}
    \vspace{-1mm}
    \item \parbox[t]{\dimexpr\textwidth-\leftmargin}{We demonstrate the superiority of our method in terms of classification and semantic recovery on four benchmark datasets by comparing with five competitive baseline methods. We also provide rich in-depth explorations of our method for unseen category discovery.}
    \vspace{-1mm}
\end{itemize}

\vspace{-2mm}
\section{Related Works}

\noindent\textbf{Generalized Zero-Shot Learning.} The demand to recognize novel classes not present during training without additional labeling motivates the area of Zero-Shot Learning (ZSL). In ZSL, class-level semantic attributes are provided to aid knowledge transfer from source to the target dataset. Later, due to the disjoint nature of the source and target label spaces, GZSL is hence proposed~\cite{pourpanah2022review}, where the target dataset consists of categories present in the source dataset and additional novel categories. Existing GZSL methods can be categorized into embedding-based and generative-based methods. Embedding-based methods learn to map visual features to the semantic space and make predictions based on the similarity between predicted semantics and ground truth \cite{frome2013devise, akata2015sje, romera2015eszsl, xu2020attribute, xie2019attentive, xie2020region, liu2021goal}. Generative-based methods synthesize samples from semantic attributes through Generative Adversarial Networks (GANs) \cite{goodfellow2020gan} or Variational Autoencoders (VAEs) \cite{kingma2013vae}, and reformulate the GZSL problem to a conventional supervised learning problem by training a classifier with generated samples \cite{li2019leveraging, xian2018feature, mishra2018generative, zhu2018generative, yu2020episode}.

\noindent\textbf{Open-Set Recognition.} OSR extends the traditional closed-set classification setting by assuming the target samples contain categories seen during training and additional unseen categories. Under OSR, classifiers need to be able to recognize seen samples and reject unknown samples \cite{geng2020osr}. OSR methods usually adopt a threshold-based strategy, where the threshold can be empirically chosen. Existing methods differ in how they discriminate seen categories, including SVM-based \cite{cortes1995svm, scheirer2012toward, scheirer2014probability, jain2014multi}, distance-based \cite{bendale2015towards, papa2009supervised, mendes2017nearest}, and deep learning based methods\cite{bendale2016towards, hassen2020learning, shu2017doc, hughes2017medical, kardan2017mitigating}. 

In addition to GZSL and OSR, we notice the recently proposed Semantic Recovery Open-Set Domain Adaptation (SR-OSDA) \cite{jing2021taotao} and the Generalized Category Discovery (GCD) \cite{vaze2022generalized, wen2022simple, chiaroni2022mutual} are related to our setting. SR-OSDA aims to recognize both seen and unseen samples in the target dataset and recover their semantic attributes. 
The goal of GCD is to automatically categorize a partially labeled dataset of images, where the unlabeled images contain both seen and novel categories, but GCD does not consider semantic recovery.

\section{Motivation and Problem Definition}\label{sec:problem}
GZSL aims to recognize categories both seen and unseen during training, given semantic attributes of all classes. However, novel categories are pre-defined with semantic labels and are known to the model, which can be impractical for some applications. OSR aims to detect unseen samples without prior information but treats them as one group, failing to explore their innate structures further. Moreover, the semantic attributes of visual categories are not utilized under OSR, which prevents in-depth analyses and interpretations of the structures of novel categories.

Due to the limitations of GZSL and OSR, we consider the  \textbf{Zero-Knowledge Zero-Shot Learning (ZK-ZSL)} problem, which focuses on recognizing seen and unseen categories without prior knowledge of novel semantic labels while uncovering the partitional structures of novel categories and their missing semantic attributes. Inspired by the paradigm style in \citet{pourpanah2022review}, we illustrate the main differences among GZSL, OSR, and ZK-ZSL in Figure~\ref{fig:schemaicviews}. Compared with GZSL and OSR, ZK-ZSL is more realistic in that it does not assume prior knowledge of the unseen categories, and it can support richer analysis and interpretation of the discovered novel categories. Here we provide problem formulation of ZK-ZSL: Let $D_s = \{X_s, Y_s, A_s\}$ be the source dataset, and $D_t = \{X_t\}$ be the target dataset. Given both $D_s$ and $D_t$, ZK-ZSL aims to find $X_t$'s corresponding $Y_t$, which contains all visual categories in $Y_s$ and some additional unseen categories $Y_u = Y_t \setminus Y_s$, and recover their semantic attributes $A_t$. Table~\ref{tab:notation} summarizes the major notations used in this paper.\footnote{In the following, we might put $s/t$ in the superscript and the sample index in the subscript if needed.}

\begin{wraptable}{R}{0.5\textwidth}
\vspace{-4mm}
\centering
\caption{Notions and descriptions}\label{tab:notation}
\resizebox{.9 \linewidth}{!}{
    \begin{tabular}{lll}
    \toprule
    Notion & Type & Description \\
    \midrule
    $X_s/X_t$ & Input & source/target visual features \\ 
     $A_s/A_t$ & Input & source/target semantic attributes \\ 
     $Y_s/Y_t$ & Input & source/target labels \\ 
     $K_s/K_t$ & Input & source/target number of categories \\
     $N_s/N_t$ & Input & source/target number of instances \\
     $d$ & Input & dimension of semantic attribute \\
     \hline
     $z_s / z_t$ & Learnable & source/target embedding \\
     $\hat{a}_s / \hat{a}_t$ & Learnable & source/target predicted semantic \\
     $\Tilde{\mu}_{k}$ & Learnable & seen class prototype by pseudo label \\
     $\mu_{k}$ & Learnable & class prototype \\
    \bottomrule
    \end{tabular} \vspace{-10mm}
    }
\end{wraptable}

The challenges associated with ZK-ZSL are twofold. First, the model needs to provide partitional structures of unseen categories. However, training a visual classifier only with labeled seen samples can only help distinguish between seen/unseen categories but cannot further separate the fine-grained categories in the unseen class. In addition, due to mismatched label spaces in the source and target dataset, learning such a classifier could destroy the clustering structures of novel categories. Second, since the semantic attributes of unseen categories are unavailable, learning only with seen semantics is prone to overfitting and results in biased semantic prediction.

\section{Proposed Method}
In this section, we introduce our framework for ZK-ZSL and the objective function to train our framework. 

\subsection{Framework Overview}
\begin{figure}[tp]
    \centering
    \resizebox{.65 \linewidth}{!}{
    \includegraphics[width=\linewidth]{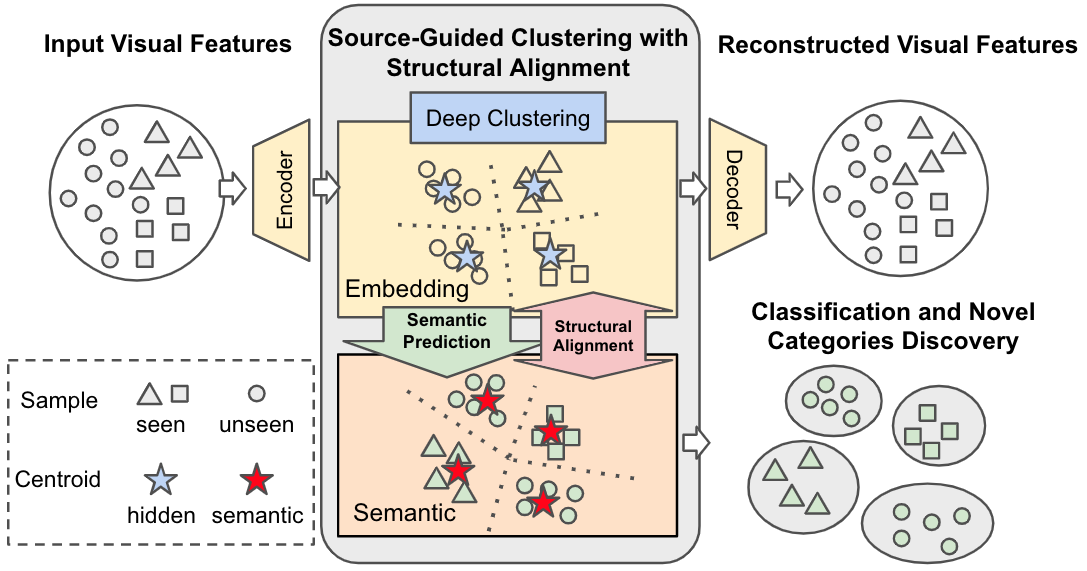}
    }
    \caption{Overview of our proposed framework. The deep clustering component guided by the source class structures uncovers the clustering structures of the target dataset. The semantic prediction component maps the embedding space to the semantic space. The structural alignment component supports semantic learning of unseen categories with recovered structures.    }
    \label{fig:framework}
    \vspace{-3mm}
\end{figure}

Figure~\ref{fig:framework} shows our proposed framework for ZK-ZSL. In order to discover and differentiate novel categories while recognizing seen classes, our method learns to recover the clustering structures of both seen and unseen categories in the target dataset, where the learning of seen class structures is guided by the source labels. In addition, to mitigate the overfitting brought about by the unavailability of unseen semantic labels, we apply a structural alignment loss to support semantic learning with the recovered structures of unseen categories. Therefore, our framework consists of three main components for source-guided clustering, semantic prediction, and structural alignment. 
 The source-guided clustering component learns to recover the clustering structures of both seen and unseen categories on a hidden embedding space where the learning of seen class structures is supported by labeled source data. The semantic prediction component maps the hidden embedding to the semantic space. The structural alignment component guides the semantic learning of novel categories with their recovered clustering structures.

\subsection{Objective Function}
Our learning objective consists of three main components: source-guided clustering loss, semantic prediction loss, and structural alignment loss. We denote our encoder, decoder, and semantic predictor as $g: x \rightarrow z$, $h: z \rightarrow x$, and $f: z \rightarrow a$, where $x$, $z$, and $a$ represent visual feature, hidden embedding, and semantic attribute spaces, respectively, and we define learnable cluster centroids as $\mathcal{C} = \{ \mu_{1}, ..., \mu_{K_t} \}$.

\noindent\textbf{Source-guided Clustering Loss.} Our method learns to recover clustering structures of both seen and unseen categories guided by source labels. To achieve this, we utilize an auto-encoder to extract hidden embeddings, apply clustering regularization on the embedding space to separate both seen and unseen categories, and align seen cluster centroids with source labels. In this way, we break down the source-guided clustering loss into three parts: {{self-reconstruction}}, {{clustering regularization}}, and {{source centroid alignment}}.

\noindent{\textit{Self-reconstruction.}} Self-reconstruction loss helps the encoder and decoder extract clustering-favorable hidden embeddings with minimal information loss. It is defined as:
\begin{equation}
    \mathcal{L}_{self} = \frac{1}{N_s} \sum_i^{N_s} \lVert x^s_i - h \circ g(x^s_i) \rVert^2_2 + \frac{1}{N_t} \sum_i^{N_t} \lVert x^t_i - h \circ g(x^t_i) \rVert^2_2, 
\end{equation} where $\circ$ represents function composition. 

\noindent{\textit{Clustering Regularization.}} Following Deep Embedding Clustering~\cite{xie2016unsupervised}, we model the cluster assignment with a probability distribution. Specifically, the similarity between an embedding instance $z_i$ to a cluster centroid $\mu_k$ is measured with a Student-t distribution:
\begin{equation}
p^{z}_{ik} = \frac{(1 + \lVert z_i - {\mu}_k \rVert^{2}_2)^{-1}}{\sum_{j=1}^{K_t}(1 + \lVert z_i - {\mu}_j \rVert^{2}_2)^{-1}}.
\label{eq:cluster}
\end{equation}
Subsequently, an auxiliary target distribution is employed to strengthen cluster prediction. The target distribution is defined by
\begin{equation}
q^{z}_{ik} = \frac{(p^{z}_{ik})^2 / \sum_j^{N_t} p^z_{jk}}{\sum_{j'}^{K_t} (p^z_{ij'})^2 / \sum_{j}^{N_t} p^z_{jj'}}.
\end{equation}
The clustering regularization loss measures the difference between the cluster assignment distribution and target distribution. By minimizing their divergence, data points are pulled toward the likely cluster centroid and pushed away from others. Mathematically, the clustering regularization loss is defined as
\begin{equation}
    \mathcal{L}_{reg} = \frac{1}{N_t} \sum_{i=1}^{N_t} D_{KL}(p^{z}_i || q^{z}_i) = \frac{1}{N_t} \sum_{i=1}^{N_t} \sum_{k=1}^{K_t} p^z_{ik} \log \frac{p^z_{ik}}{q^z_{ik}}.
\end{equation}

\noindent{\textit{Source Centroid Alignment.}} The source centroid alignment loss pulls source samples toward their corresponding cluster centroids. This is achieved by optimizing the prototypical probability distribution defined by the distance between data points and cluster centroids. The prototypical probability of a data point $z_i$ given a cluster centroid $\mu_k$ is  
\begin{equation}
\label{eq:prototypical}
\mathcal{P}(z_i | \mu_k) = \frac{\exp(- d(z_i, \mu_k ))}{\sum_{k' \leq K_s} \exp(- d(z_i, \mu_{k'}))},   
\end{equation}
where $d$ is an Euclidean distance function and the source centroid alignment loss is
\begin{equation}
\mathcal{L}_{cent} = \frac{1}{N_s} \sum_{i=1}^{N_s}\mathcal{L}(\mathcal{P}_i, y_i),    
\end{equation}
where $\mathcal{L}$ denotes the cross entropy loss. To address the potential distribution drift of seen categories between the source and target dataset, we incorporate an optional loss to reduce the gap between source and target seen class centroids. We first define the expected target seen class centroid 
\begin{equation}
    \Tilde{\mu}_k := \mathbb{E}_{\hat{y}^t_i = k} z^t_i,
\end{equation} 
where $\hat{y}^t_i = \arg_k \max p^{z}_{ik}$ is the pseudo label obtained from cluster assignment probability. The source centroid alignment loss under distribution shift is defined as
\begin{equation}
\mathcal{L}_{cent} = \frac{1}{N_s} \sum_{i=1}^{N_s}\mathcal{L}(\mathcal{P}_i, y^i_s) + \frac{1}{K_s} \sum_{k=1}^{K_s} \lVert \Tilde{\mu}_k - \mu_k \rVert^2_2.
\end{equation}

Finally, the source-guided clustering loss combines self-reconstruction loss, clustering regularization loss, and source centroid alignment loss:
\begin{equation}
    \mathcal{L}_{c} = \mathcal{L}_{self} + \mathcal{L}_{reg} + \mathcal{L}_{cent}.
\end{equation}

\noindent\textbf{Semantic Prediction Loss}. Following \citet{frome2013devise}, we use a pairwise ranking loss to learn the embedding-to-semantic mapping, where we stop penalizing false terms after the maximum margin of 0.5 is reached.
\begin{equation}
    \mathcal{L}_a = \frac{1}{N_s} \sum_{i=1}^{N_s} \sum_{k \neq y_s^i} \max \{0, 0.5 - A_{y_i}^\top a^s_i + A_k^\top a^s_i \},
\end{equation}
where $A_{k}$ is the ground truth semantic label of class $k$. 

\noindent\textbf{Structural Alignment Loss}. Structural alignment guides semantic learning by transferring the structural knowledge in the embedding space to the semantic space. Specifically, the structural alignment loss minimizes the divergence between embedding and semantic cluster structures modeled by the cluster assignment probability. We have defined the cluster assignment probability in the embedding space $p_{ik}^z$ in Eq.~\eqref{eq:cluster}. Similarly, the cluster assignment probability in the semantic space can be defined by $p^{a}_{ik}$, where
\begin{equation}
p^{a}_{ik} = \frac{(1 + \lVert a_i - f(\mu_k) \rVert^{2}_2)^{-1}}{\sum_{j=1}^{K_t}(1 + \lVert a_i - f(\mu_j) \rVert^{2}_2)^{-1}}.
\label{eq:semanticcluster}
\end{equation}
Based on Eq.~\eqref{eq:cluster} and Eq.~\eqref{eq:semanticcluster}, we define the structural alignment loss as
\begin{equation}
\mathcal{L}_{align} = \frac{1}{N_t} \sum_{i=1}^{N_t} \sum_{j=1}^{N_t}  ( {p_i^z}^\top p_j^z - {p_i^a}^{\top} p_j^a )^2,
\end{equation}
where ${p_i^z}^\top p_j^z$ measures the probability that data points $i$ and $j$ in the embedding space belonging to the same cluster.

\noindent\textbf{Overall Objective Function}. The overall learning objective is to minimize the source-guided clustering loss, semantic prediction loss, and structural alignment loss:
\begin{equation}
\min_{g, h, f, \mathcal{C}} \mathcal{L}_{c} + \alpha \mathcal{L}_a + \beta \mathcal{L}_{align},
\end{equation} where $\alpha$ and $\beta$ are trade-off parameters.

\label{sec:method}

\section{Experiments}

\subsection{Experimental Settings}

\begin{wraptable}{R}{0.5\textwidth}
\centering
\caption{
Characteristics of four benchmark datasets.
}
\scriptsize
\resizebox{.9 \linewidth}{!}{
    \begin{tabular}{c|cccccc}
    \toprule
    Dataset & Source & Target\_s & Target\_t  & $K_s$ & $K_t$ & $d$ \\
    \midrule
    APY & 5932 & 7924 & 1483 & 20 & 32 & 64 \\
    CUB & 8855 & 2973 & 1764 & 150 & 200 & 312 \\
    AWA2 & 19832 & 5685 & 4958 & 40 & 50 & 85 \\
    SUN & 12900 & 1440 & 2580 & 645 & 717 & 102 \\
    \bottomrule
    \end{tabular}
    }
        \begin{tablenotes}
   \item \parbox[t]{7.5 cm}{Note: Target\_s and Target\_t denote the sample numbers of seen and unseen categories in the target domain.}
\end{tablenotes}
\label{tab:characteristic}
 \vspace{-3mm}
\end{wraptable}

\noindent\textbf{Datasets}. We choose four commonly used attribute datasets: Attribute Pascal and Yahoo (APY) \cite{farhadi2009apy}, Animals with Attributes 2 (AWA2) \cite{lampert2013awa}, Caltech-UCSD-Birds 200-2011 (CUB) \cite{wah2011cub}, and SUN dataset \cite{xiao2010sun}. We split the data into the source and target sets by following the split strategy in \citet{xian2017benchmark}. All visual features are obtained from ResNet-101 \cite{he2016resnet} pre-trained on ImageNet \cite{deng2009imagenet}. Table \ref{tab:characteristic} summarizes the key characteristics of these datasets. Only the semantic attributes of seen categories are available for training. We use unit vectors as semantic representations. 

\noindent\textbf{Competitive Methods}. We compare our method with established methods in Generalized Zero-Shot Learning and Semantic Recovery Open-Set Domain Adaptation. For GZSL based methods, it is noteworthy that not all established methods work under our setting, where unseen semantics are not provided. For example, generative-based methods, which require unseen semantics to synthesize samples of novel categories, are not applicable under ZK-ZSL. We compare with DEVISE \cite{frome2013devise}, ESZSL \cite{romera2015eszsl}, SJE \cite{akata2015sje} and SDGZSL \cite{chen2021sdgzsl}. In particular, DEVISE, SJE, and ESZSL aim to find a mapping from visual space to semantic space, while SDGZSL learns a hidden embedding from visual features that can be decomposed to semantic-consistent representation and semantic-unrelated representation. In the next section, we elaborate on how to tailor the above methods to the ZK-ZSL setting. In addition, we compare with \citet{jing2021taotao} as the only existing SR-OSDA-based method. Their approach progressively separates seen and unseen samples with pseudo labels obtained from prototypical probabilities and utilizes a graph neural network for semantic attribute prediction to avoid overfitting. 

\noindent\textbf{Implementation}. Here we first introduce how we extend the competitive methods for novel category prediction, then elaborate on the implementation details of our method.

Since the competitive methods only predict semantic attributes or hidden embeddings, we need to extend their approaches to support novel categories prediction. To this end, we adopt a two-step procedure that first recognizes seen and unseen samples and then further separates unseen samples with K-means. We rely on the prototypical probability defined in Eq.~\eqref{eq:prototypical} for seen/unseen separation. We first define the highest prototypical probability that a data point belongs to a seen category:
\begin{equation}
    \mathcal{P}^{max}_i = \max_k \mathcal{P}(z_t^i | \mu_k).
\end{equation}
Data points with higher $\mathcal{P}^{max}$ are more likely from a seen category, while unseen samples tend to have more even prototypical probability distributions, and thus lower $\mathcal{P}^{max}$. We follow \citet{jing2021taotao} to choose a threshold where the data points with $\mathcal{P}^{max}$ higher than $\tau = \frac{1}{N_t} \sum_i \mathcal{P}^{max}_i$ are recognized as seen samples, and classified by their corresponding class centroids; those with lower $\mathcal{P}^{max}$ are regarded as unseen samples and predicted by K-means implemented by Scikit-learn \cite{scikit-learn}. Let unseen class centroids predicted by K-means be $\{ \hat{\mu}_{K_s + 1}, ..., \hat{\mu}_{K_t} \}$, we define classification prediction as 
\begin{equation}
    \hat{y}_t^i =
    \begin{cases}
      \arg \max_{k \leq K_s} \mathcal{P}(z_t^i | \mu_k) & \text{if \ } \mathcal{P}^{max}_i \geq \tau \text{,} \\
      \arg \min_{K_s < k \leq K_t} d(z_t^i, \hat{\mu}_k) & \text{otherwise.}
    \end{cases}  
\end{equation}

In our framework, the encoder and decoder constitute a de-noising autoencoder. Specifically, the encoder and decoder are 5 linear layers followed by batch normalization and Leaky ReLU activation. Their neural structures are dropout-2048-512-256-256-4096-$h$ and dropout-$h$-4096-256-256-512-2048, respectively, where $h$ is the dimension of the hidden embeddings. We set the dropout rate to $0.01$ for all experiments and choose $h = 256$ for APY and AWA2 datasets and $h = 1024$ for CUB and SUN datasets. The semantic prediction head is a linear layer with batch normalization. Our model parameters are initialized by pretraining with self-reconstruction loss and semantic prediction loss, and class centers are initialized with K-means. We use Adam optimizer \cite{kingma2014adam} with a learning rate of $1e^{-3}$ and weight decay of $1e^{-5}$ in the pretraining stage, and then learning rate is adjusted to $1e^{-4}$ for training. Learning rates are reduced by a factor of 0.1 every 200 epochs for both stages. 

\noindent\textbf{Metrics}. We evaluate model performance from two perspectives: classification and semantic recovery accuracy. 
\begin{table*}[t]
\centering
\caption{
Classification and Semantic Recovery Accuracy on four benchmark datasets. 
}
\scriptsize
\begin{tabular}{c|c|S[table-format=2.1]S[table-format=2.1]S[table-format=2.1]|S[table-format=2.1]S[table-format=2.1]S[table-format=2.1]|S[table-format=2.1]S[table-format=2.1]S[table-format=2.1]|S[table-format=2.1]S[table-format=2.1]S[table-format=2.1]}
\toprule
& \multicolumn{1}{c}{Dataset} & \multicolumn{3}{c}{APY} & \multicolumn{3}{c}{CUB} & \multicolumn{3}{c}{AWA2} & \multicolumn{3}{c}{SUN}\\
\midrule
& \multicolumn{1}{c|}{Method} & \multicolumn{1}{c}{$Acc_s$} & \multicolumn{1}{c}{$Acc_u$} & \multicolumn{1}{c|}{$Acc_h$} & \multicolumn{1}{c}{$Acc_s$} & \multicolumn{1}{c}{$Acc_u$} & \multicolumn{1}{c|}{$Acc_h$} & \multicolumn{1}{c}{$Acc_s$} & \multicolumn{1}{c}{$Acc_u$} & \multicolumn{1}{c|}{$Acc_h$} & \multicolumn{1}{c}{$Acc_s$} & \multicolumn{1}{c}{$Acc_u$} & \multicolumn{1}{c}{$Acc_h$} \\
\cmidrule{2-14}
\multirow{6}{*}{\vspace{-10mm}\turnbox{90}{Classification}} & DEVISE$^*$ & \BlueText 76.4 & 13.0 & 22.2 & 37.1 & 15.8 & 22.2 & 69.2 & 24.5 & 36.2 & 18.4 & 21.9 & 20.0 \\
 & ESZSL$^*$ & \RedText 77.2 & 17.0 & 27.8 & 35.9 & 23.8 & 28.6 & 67.5 & \BlueText 40.3 & \BlueText 50.4 & 16.6 & 18.6 & 17.6 \\
 & SJE$^*$ & 74.1 & 9.2 & 16.3 & 34.1 & 9.3 & 14.6 & 57.0 & 13.2 & 21.5 & 17.2 & 17.6 & 17.4 \\
 & SDGZSL$^\diamond$ & 71.1 & \BlueText 22.5 & \BlueText 34.2 & \RedText 58.0 & 33.6 & \RedText 42.5 & \BlueText 79.6 & 27.9 & 41.4 & \RedText 32.8 & \RedText 26.5 & \RedText 29.3 \\
 & SR-OSDA$^*$ & 24.2 & 15.4 & 18.8 & 2.7 & 11.7 & 4.4 & 27.0 & 18.6 & 22.0 & 5.0 & 11.7 & 9.7 \\
\cmidrule{2-14}
 & Ours & 48.2 & \RedText 30.6 & \RedText 37.4 & \BlueText 43.8 & \RedText 35.4 & \BlueText 39.2 & \RedText 82.8 & \RedText 47.6 & \RedText 60.5 & \BlueText 23.5 & \BlueText 24.8 & \BlueText 24.1 \\
 \midrule
 & \multicolumn{1}{c|}{Method} & \multicolumn{1}{c}{$SR_s$} & \multicolumn{1}{c}{$SR_u$} & \multicolumn{1}{c|}{$SR_h$} & \multicolumn{1}{c}{$SR_s$} & \multicolumn{1}{c}{$SR_u$} & \multicolumn{1}{c|}{$SR_h$} & \multicolumn{1}{c}{$SR_s$} & \multicolumn{1}{c}{$SR_u$} & \multicolumn{1}{c|}{$SR_h$} & \multicolumn{1}{c}{$SR_s$} & \multicolumn{1}{c}{$SR_u$} & \multicolumn{1}{c}{$SR_h$} \\
\cmidrule{2-14}
 \multirow{6}{*}{\vspace{-12mm}\turnbox{90}{Semantic Recovery}} & DEVISE & 2.6 & \BlueText 9.7 & 4.1 & 37.8 & \BlueText 24.4 & \BlueText 29.6 & 61.1 & \BlueText 26.7 & \BlueText 37.2 & \BlueText 29.4 & 14.8 & 19.7 \\
 & ESZSL & \RedText 83.6 & 1.4 & 2.8 & \RedText 55.5 & 11.5 & 19.0 & \RedText 88.5 & 2.2 & 4.3 & 27.3 & 13.1 & 17.7 \\
 & SJE & 7.5 & 9.4 & \BlueText 8.3 & 0.8 & 11.6 & 1.4 & 41.8 & 20.4 & 27.4 & 27.0 & \BlueText 16.8 & \BlueText 20.7 \\
 & SR-OSDA & 3.4 & 0.7 & 1.2 & 0.3 & 0.2 & 0.3 & 2.7 & 0.4 & 0.7 & 0.2 & 0.3 & 0.2 \\
\cmidrule{2-14}
& Ours & \BlueText 60.4 & \RedText 17.5 & \RedText 27.2 & \BlueText 52.4 & \RedText 25.1 & \RedText 33.9 & \BlueText 84.6 & \RedText 32.9 & \RedText 47.4 & \RedText 36.8 & \RedText 18.2 & \RedText 24.4 \\
\bottomrule 
\end{tabular}
\begin{tablenotes}
   \item \hspace{12mm} Note: $*$ represents evaluation in semantic space, and $\diamond$ represents evaluation in embedding space. 
\end{tablenotes}
\vspace{-2mm}
\label{tab:Acc}
\end{table*}

In the same spirit of the evaluation metrics proposed in \citet{xian2017benchmark}, our classification accuracy measures the class average accuracy of seen and unseen categories and their harmonic mean. We define the classification accuracy of seen categories as the average per-class accuracy:
\begin{equation}
    Acc_s = \frac{1}{N_s} \sum_{k \leq K_s} \frac{\sum_i \mathds{1}_{\hat{y}_t^i = y_t^i}}{\sum_i  \mathds{1}_{y_t^i = k}},
\end{equation} where $\mathds{1}$ is an indicator function. Similarly, we define the classification accuracy of unseen category as
\begin{equation}
    Acc_u = \frac{1}{N_t - N_s} \sum_{K_s < k \leq K_t} \frac{\sum_i \mathds{1}_{map(\hat{y}_t^i) = y_t^i}}{\sum_i  \mathds{1}_{y_t^i = k}},
\end{equation} where $map()$ is a permutation function that maps predicted unseen categories to ground-truth unseen counterparts. Finally, we evaluate the harmonic mean of seen and unseen classification accuracy:
\begin{equation}
    Acc_h = 2 \times \frac{Acc_s \times Acc_u}{Acc_s + Acc_u}.
\end{equation}

\begin{figure*}[t]
    \centering
    \begin{subfigure}{0.2 \textwidth}
        \resizebox{1 \textwidth}{!}{{\includegraphics{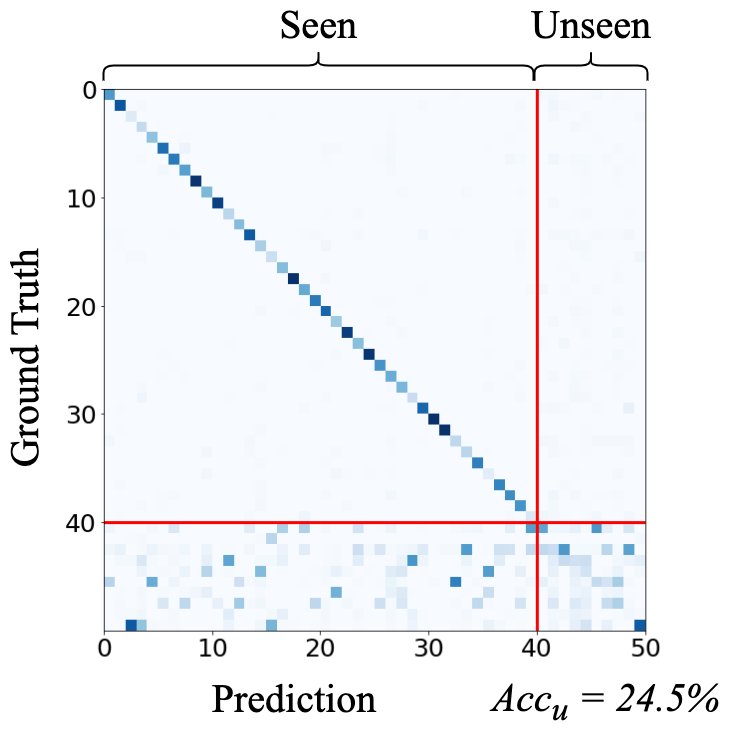}};
        }
    \caption{DEVISE}
    \label{fig:confuse_a}
    \end{subfigure}
    \begin{subfigure}{0.2 \textwidth}
        \resizebox{\textwidth}{!}{{\includegraphics{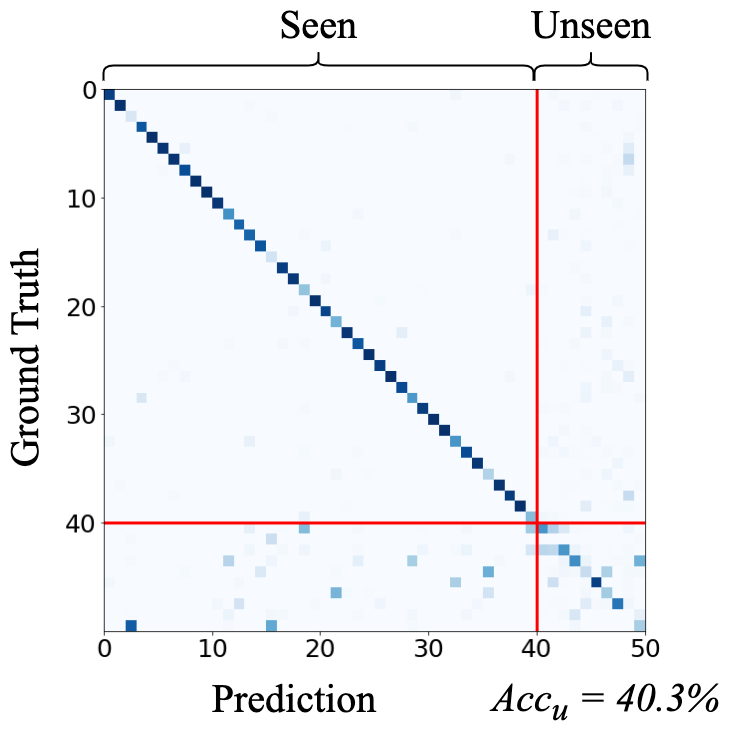}};
        }
    \caption{ESZSL}
    \label{fig:confuse_b}
    \end{subfigure}
    \begin{subfigure}{0.2 \textwidth}
        \resizebox{\textwidth}{!}{{\includegraphics{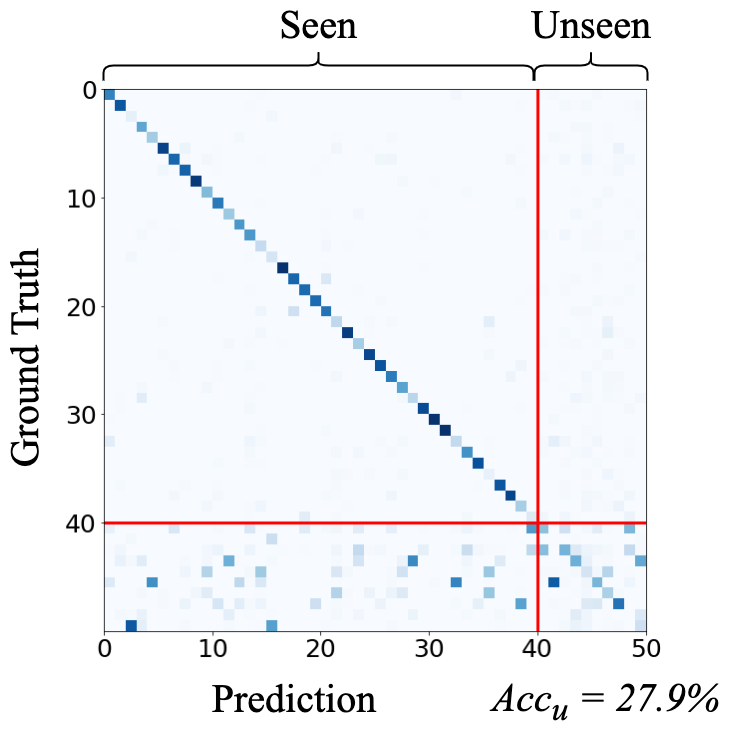}};
        }
    \caption{SDGZSL}
    \label{fig:confuse_b}
    \end{subfigure}
    \begin{subfigure}{0.38 \textwidth}
        \resizebox{\textwidth}{!}{{\includegraphics{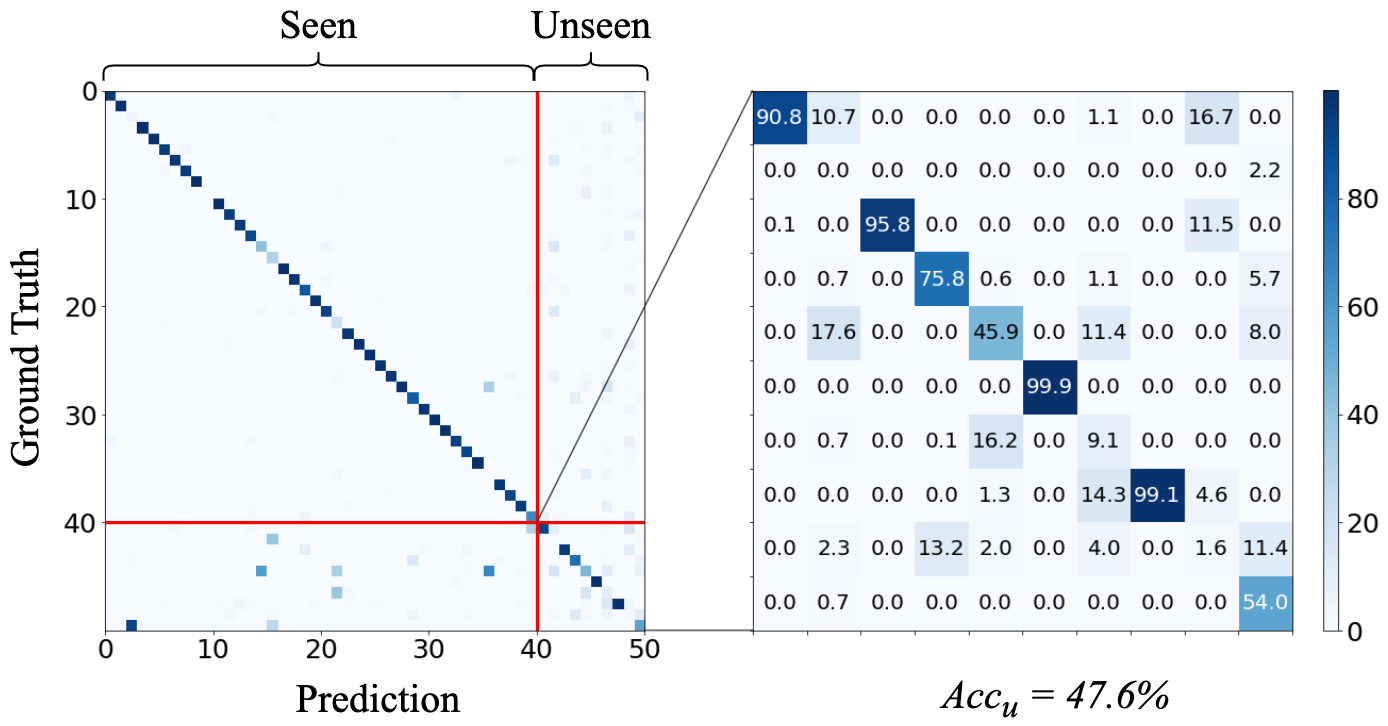}};
        }
    \caption{Ours}
    \label{fig:confuse_c}
    \end{subfigure}\vspace{-1mm}
    \caption{Confusion matrices of seen and unseen categories on the AWA2 dataset. Recommended to zoom in for better visualization.}\vspace{-3mm}
    \label{confusion matrices}
\end{figure*}

In addition to classification accuracy, we want to evaluate the model's ability to accurately uncover the target dataset's semantic attributes. Following \citet{xian2017benchmark}, we define the semantic recovery prediction as
\begin{equation}
    \Tilde{y}_t^i = \arg \max_k A_k^\top a_t^i.
\end{equation} 
Similarly, semantic recovery accuracies $SR_s$, $SR_u$, and $SR_h$ follow the derivation of classification accuracies. 

For DEVISE, SJE, and ESZSL, we evaluate their classification accuracies with their predicted semantic attributes, where the class centroids of seen categories are ground truth semantic labels. We evaluate SDGZSL classification ability with its generated semantic-consistent representation, and because it does not have a semantic prediction component, we exclude it from semantic recovery evaluation. We use the cosine distance when calculating prototypical probability in the semantic space and use Euclidean distance in the embedding space.

\subsection{Performance}

\begin{figure*}[t]
    \centering
    \subcaptionbox{\label{sfig:a} Bobcat}{\includegraphics[width=3.5cm]{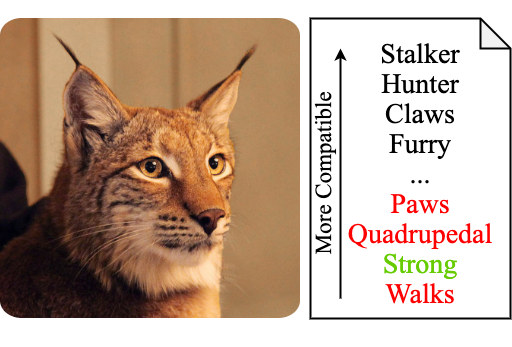}}
    \hspace*{0.1cm}
    \subcaptionbox{\label{sfig:b} Dolphin}{\includegraphics[width=3.5cm]{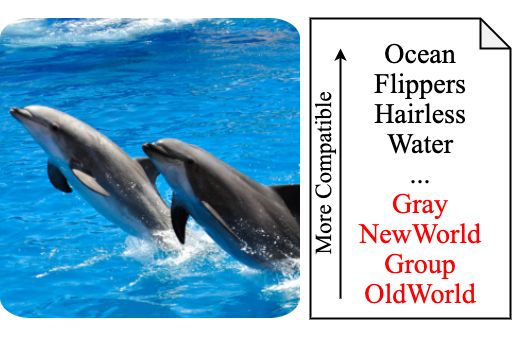}}
    \hspace*{0.1cm}
    \subcaptionbox{\label{sfig:c} Sheep}{\includegraphics[width=3.5cm]{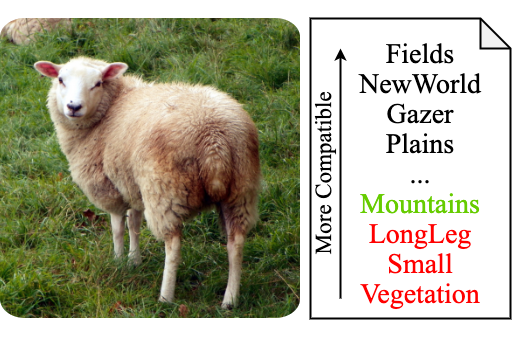}}
    \hspace*{0.1cm}
    \subcaptionbox{\label{sfig:d} Giraffe}{\includegraphics[width=3.5cm]{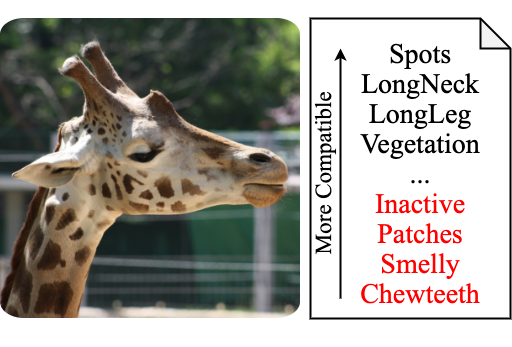}}
    \subcaptionbox{\label{sfig:e} Rat}{\includegraphics[width=3.5cm]{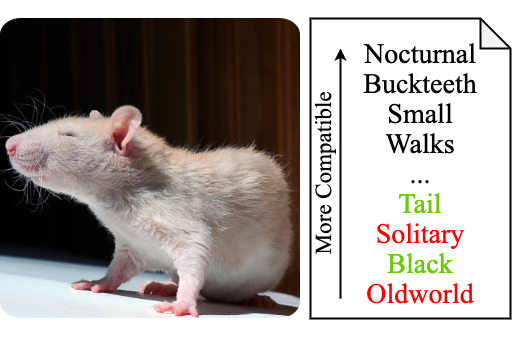}}
    \hspace*{0.1cm}
    \subcaptionbox{\label{sfig:f} Seal}{\includegraphics[width=3.5cm]{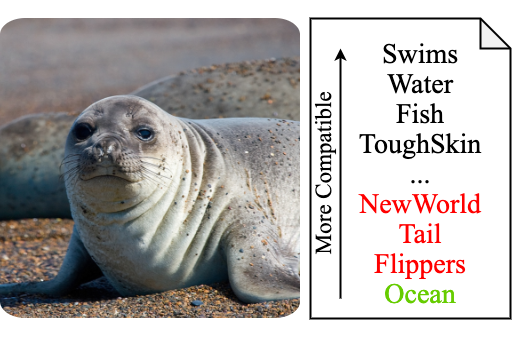}}
    \hspace*{0.1cm}
    \subcaptionbox{\label{sfig:g} Bat}{\includegraphics[width=3.5cm]{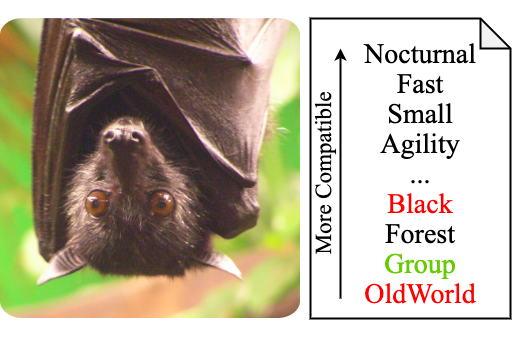}}
    \hspace*{0.1cm}
    \subcaptionbox{\label{sfig:d} Horse}
    {\includegraphics[width=3.5cm]{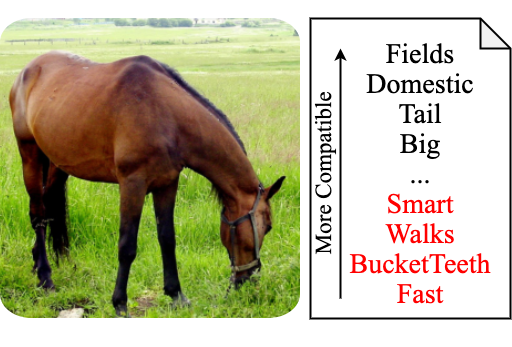}}\vspace{-2mm}
    \caption{Examples of novel categories and predicted semantic attributes. The predicted semantics are ordered by compatibility with ground-truth labels, where the top is the most compatible semantic attribute. \textcolor{red}{Red} indicates a wrong prediction, and \textcolor{green}{Green} indicates a wrong prediction but is consistent with the given instance. }
    \vspace{-2mm}
    \label{fig:examples}
\end{figure*}

\begin{figure*}[t]
    \centering
    \subcaptionbox{ \label{sfig:a} ResNet}
    {\vspace{-2mm} \includegraphics[width=3.5cm]
    {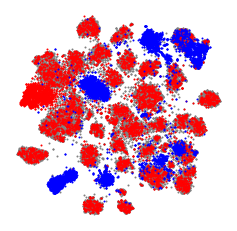}}
    \subcaptionbox{
        \label{sfig:b} SJE}{\vspace{-2mm} \includegraphics[width=3.5cm]{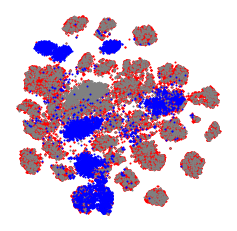}
    }
    \subcaptionbox{\label{sfig:c} Ours (embedding)}{\vspace{-2mm} \includegraphics[width=3.5cm]{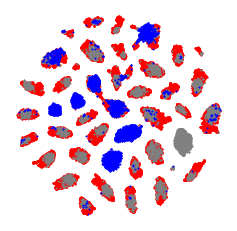}}
    \subcaptionbox{\label{sfig:d} Ours (semantic)}{\vspace{-2mm} \includegraphics[width=3.5cm]{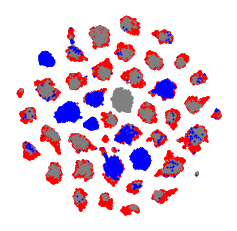}}
    \vspace{-2mm}
    \caption{t-SNE visualization of (a) ResNet features, (b) semantic attribute predicted by SJE \cite{akata2015sje}, (c) hidden embedding generated by our method, and (d) semantic attribute predicted by our method. \textcolor{red}{Red} represents source data points. \textcolor{blue}{Blue} and \textcolor{gray}{Gray} highlight data points in the target dataset that belong to unseen and seen categories, respectively.}
    \label{fig:visualization}
    \vspace{-2mm}
\end{figure*}

Table \ref{tab:Acc} summarizes the classification accuracy and semantic recovery accuracy for four benchmark datasets, where red and blue highlight the highest and second highest accuracies. 

For the classification accuracy measurement, we obverse that our method attains the best overall performance in APY and AWA2 while achieving the second-highest overall accuracy in CUB and SUN. It is noteworthy that for the AWA2 dataset, our method brings a 10.1\% overall accuracy improvement compared with the second-highest baseline. Our superior performance in APY and AWA2 partly results from our improved unseen category classification accuracy, which demonstrates the effectiveness of our structural alignment and source-guided clustering component. In comparison, we see a trend of overfitting on the source dataset for some baseline methods, which destroys the clustering structure of unseen samples. For example, regarding the performance on the APY dataset, DEVISE achieves 76.4\% seen classification accuracy but only attains 13.0\% for the unseen category. In addition, we observe our method does not improve overall classification accuracy in CUB and SUN, compared with the best-performing SDGZSL baseline. We hypothesize that this is due to a reduced number of samples per class in the target dataset. There are, on average, 293 and 212 samples per class in APY and AWA2, respectively, but for CUB and SUN there are only 23 and 6 samples. This leads to (1) distribution drift of seen category samples, which undermines our method's capability to classify seen samples correctly, and (2) underrepresented novel class samples, which complicates their clustering structures recovery.  

For the semantic recovery accuracy measurement, our method achieves the best performance in terms of unseen category semantic recovery and overall semantic recovery accuracy. This demonstrates that our structural alignment loss can effectively prevent overfitting on the source semantic data and accurately predict the semantic attributes of unseen categories. In comparison, we observe a similar pattern in classification that some baselines can accurately recover semantics of seen categories, but not for unseen samples. For example, ESZSL achieves 83.6\% for seen semantic recovery but only 1.4\% for unseen categories. 

\subsection{Factor Exploration}
To further explore our method, we explore confusion matrices of seen and unseen categories, visualize representations obtained by our method, showcase some representative samples and their associated predictions, analyze the ablation study of our model components, and visualize our predicted semantic attributes of novel categories with word clouds. Finally, we explore hyper-parameters $\alpha$ and $\beta$.

\noindent\textbf{Confusion Matrices.} We visualize confusion matrices of different methods on the AWA2 dataset, including DEVISE, ESZSL, SDGZSL, and ours in Figure \ref{confusion matrices}. We observe that our method has much fewer false predictions for novel categories. It is noteworthy that our method classifies four unseen categories with 90+\% accuracy. We also see that our method improves classification accuracy for seen categories.

\begin{wrapfigure}[17]{R}{0.5\textwidth}
    \centering{\includegraphics[width=8cm]{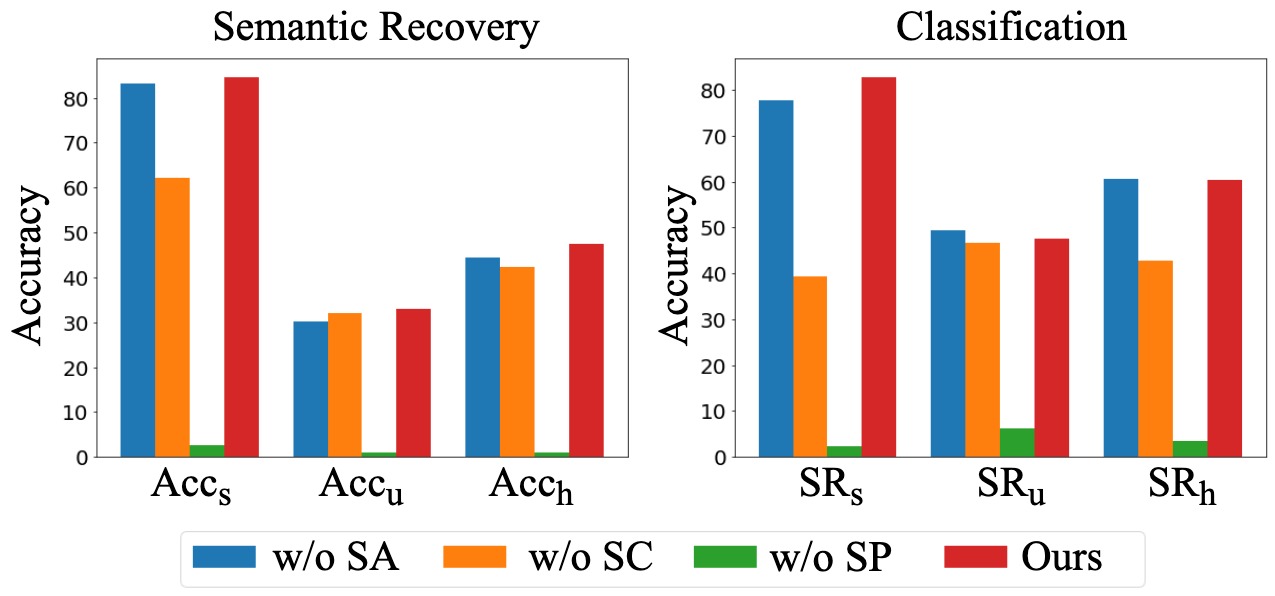}}\vspace{-2mm}
    \caption{Ablation study results of our proposed method on AWA2 dataset. We report the performance of variants that are trained without structural alignment loss (SA), without source-guided clustering loss (SC), without semantic prediction loss (SP), and with our proposed method. The left shows semantic recovery accuracy, and the right shows classification accuracy.}
    \label{fig:ablation}
    \vspace{-4mm}
\end{wrapfigure}

\begin{figure}[t]
    \centering{
    \vspace{-4mm}
    \includegraphics[width=0.95\textwidth]{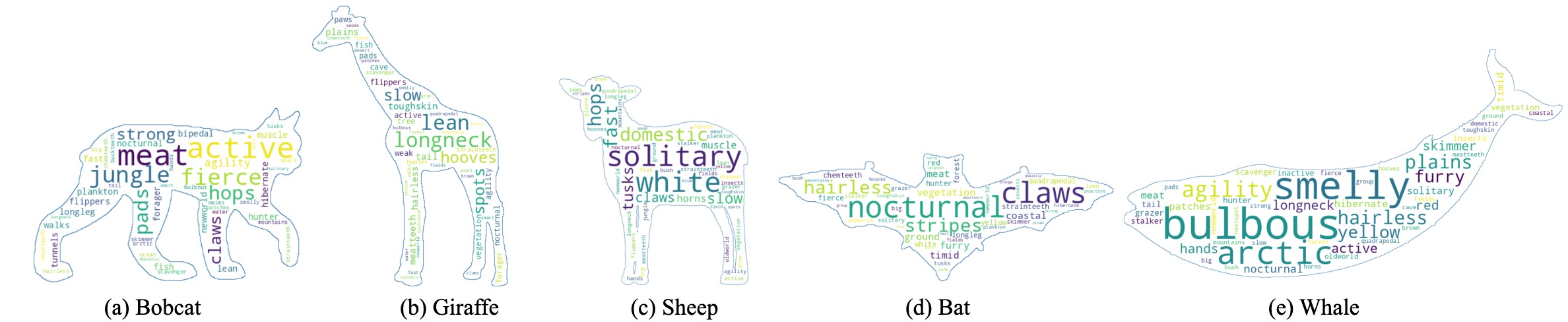}}\vspace{-1mm}
    \caption{Word clouds of novel category semantics predicted by our model, including (a) bobcat, (b) giraffe, (c) sheep, (d) bat, and (e) whale.  Recommended to zoom in for better visualization.}
    \label{fig:WordCloud}
    \vspace{-5mm}
\end{figure}

\noindent\textbf{Representation Visualization.}
In Figure \ref{fig:visualization}, we visualize different representations of the AWA2 dataset with t-SNE \cite{van2008tsne}, including the ResNet features, semantic attributes predicted by SJE, our method's hidden embedding, and our predicted semantic attributes. Red dots represent data points in the source dataset, while blue and gray dots represent target data points that belong to seen and unseen categories, respectively. For our hidden embedding visualization, we observe a much clearer separation among clusters compared with the ResNet input and SJE prediction, which is achieved with our source-guided clustering component. In addition, our recovered semantic attributes have similar structures of hidden embeddings, showing that our structural alignment loss is effective. 

\noindent\textbf{Qualitative Analysis.} In Figure \ref{fig:examples}, we showcase some representative samples from novel categories in the AWA2 dataset, including the input image and predicted semantic attributes ordered by compatibility to the ground-truth semantic labels. The compatibility is calculated by an element-wise multiplication between the predicted semantic attribute scores and ground truth. We regard a predicted attribute with positive compatibility as a correct prediction and vice versa. The most compatible predicted semantics are discriminative for the given category. For least compatible attributes, some valid attributes are not predicted, which is indicated by red in Figure \ref{fig:examples}, and some predicted attributes (in green) are not relevant to the given category but reasonable for the given instance. 

\noindent\textbf{Ablation Study.} We explore the effectiveness of our designed components with an ablation study. Specifically, we individually train variants of our framework with the AWA2 dataset that are (1) without structural alignment loss $\mathcal{L}_{align}$, (2) without source-guided clustering loss $\mathcal{L}_{c}$, and (3) without semantic prediction loss $\mathcal{L}_{a}$. In Figure \ref{fig:ablation}, we report and compare the classification accuracy and semantic recovery accuracy. We first observe that our method achieves the best overall performance when trained with all objectives. In addition, semantic recovery accuracy and seen category classification accuracy are improved with structural alignment loss, which indicates that structural alignment helps hidden embedding and semantic learning. 

\noindent\textbf{Novel Category Visualization.} Figure \ref{fig:WordCloud} visualizes our predicted semantic attributes of discovered novel categories through word clouds generated by \citet{wordcloud}. We select the 50 most significant attributes ordered by their predicted scores for visualization and adjust font sizes according to their relative ranking. We observe that the most significant attributes are consistent with the corresponding category. 

\noindent\textbf{Hyperparameter Analysis.} $\alpha$ and $\beta$ control the contributions of the semantic prediction loss and structural alignment loss and, by default, are set to 1.0 and 1.0. Here we explore the volatility of these parameters and verify our default choice by doing a hyperparameter analysis on the AWA2 dataset. When $\alpha$ is fixed at 1.0, and $\beta$ is explored in the range of $\{0, 0.1, 0.5, 1.0, 5.0, 10.0\}$, we observe that the classification and semantic recovery accuracy increase as $\beta$ increases, and reach the optimal values of $59.7\%$ and $50.9\%$ when $\alpha = 1.0$ and $\beta = 1.0$. As $\beta$ further increases, $Acc_u$ and $SR_u$ decline to $30.6\%$ and $45.4\%$. When $\beta$ is pinned at 1.0 and $\alpha$  is set in the range of $\{0, 0.1, 0.5, 1.0, 5.0, 10.0\}$, the $Acc_u$ and $SR_u$ form an inverted U pattern where they start at $47.0\%$ and $0.3\%$, reach the optimal combination, then reduce to $56.8\%$ and $42.1\%$. The optimal values are obtained at $\alpha = \beta = 1.0$.

\section{Conclusion}
In this work, we explore a novel and practical setting in visual recognition named Zero-Knowledge Zero-Shot Learning, addressing the limitations of GZSL and OSR. ZK-ZSL aims to recognize categories present in the source dataset as well as additional novel categories without prior knowledge and recover their semantic attributes. We consider the challenges associated with ZK-ZSL from two aspects: (1) the demand to further separate unseen categories, where training a classifier with only source labels could destroy the structures of unseen categories, and (2) the unavailability of unseen semantic labels that could cause inaccurate and biased semantic predictions. Our proposed method learns to recover the clustering structures of both seen and unseen categories in the target dataset, where the seen class centroids are guided by source labels. A novel structural alignment loss is designed to aid semantic learning of unseen categories with their recovered clustering structures. Experimental analyses show that our method achieves superior performance in terms of classification and semantic recovery on four benchmark datasets.

\clearpage
\bibliographystyle{plainnat}
\bibliography{references}

\end{document}